\documentclass[final,3p,times,twocolumn,number]{elsarticle}

\usepackage{amssymb}
\usepackage{amsmath}

\usepackage{svg}
\usepackage{multirow}
\usepackage{makecell}
\usepackage{url}
\usepackage{parskip}
\usepackage{hyperref}

\begin{document}

\journal{Pattern Recognition Letters}

\begin{frontmatter}



\title{On the Temporality for Sketch Representation Learning} 


\author[label1]{Marcelo Isaias de Moraes Junior \corref{cor1}} 
\ead{marcelo.junior@usp.br}
\author[label1,label2]{Moacir Antonelli Ponti} 
\ead{moacir@icmc.usp.br}

\affiliation[label1]{organization={Institute of Mathematical and Computer Sciences, University of São Paulo},
            city={São Carlos},
            state={SP},
            country={Brazil}}

\affiliation[label2]{organization={Mercado Libre},
            city={Osasco},
            state={SP},
            country={Brazil}}

\cortext[cor1]{Corresponding author}

\begin{abstract}
Sketches are simple human hand-drawn abstractions of complex scenes and real-world objects. Although the field of sketch representation learning has advanced significantly, there is still a gap in understanding the true relevance of the temporal aspect to the quality of these representations. This work investigates whether it is indeed justifiable to treat sketches as sequences, as well as which internal orders play a more relevant role. The results indicate that, although the use of traditional positional encodings is valid for modeling sketches as sequences, absolute coordinates consistently outperform relative ones. Furthermore, non-autoregressive decoders outperform their autoregressive counterparts. Finally, the importance of temporality was shown to depend on both the order considered and the task evaluated.
\end{abstract}




\end{frontmatter}

\setlength{\parskip}{0pt}


\section{Introduction}
\label{sec:introduction}

Sketch representation learning has been gaining attention, especially for being an abstract and simple human-friendly data \cite{SketchRNN}. Applications in this field range from classification, segmentation, and reconstruction tasks on existing sketches to generate new ones \cite{SketchRNN, Sketch-a-Net, sketchformer, Sketchformer++, SketchINR, StrokeCloud}. Although the raster representation was the standard choice with Convolutional Neural Networks (CNN) and Vision Transformers (ViT) based models \cite{Sketch-a-Net, SketchCLIP}, they are unable to capture sequence and stroke structure which are important features to representation learning \cite{TUBerlin, sketchformer, SketchINR, SketchGNN, Sketch-segformer}. 

In a opposite direction, vector-based operates directly on the sketch data and is divided in sequence-based and spatial-based methods \cite{SketchGNN, sketchformer}. The former advocates the use of a point/coordinate base representation referred to as Stroke-5 (as defined in section~\ref{ss:sketch}), treating sketch as a sequence of relative point coordinates that models the human drawing process \cite{SketchRNN}. Whereas the latter discards the temporality and focuses the spatial aspect of sketches, such absolute coordinates and hidden graph structures \cite{SketchGNN, Sketch-segformer, bhunia2021vectorization}. Supporting this view, some works question the importance of sketch temporality, specifically whether stroke order matters for sketch classification \cite{sketchformer}, and disregard the causal drawing sequence in sketch reconstruction and generation by adopting non-autoregressive decoding \cite{SketchINR, StrokeCloud}.

All in all, it remains unanswered for what extent and tasks sketch temporality is important for representation learning. To explore this issue more deeply, we ask the following research questions (RQ) about the sketch sequence-related treatment: 
\begin{itemize}
    \item \textbf{RQ1: Is autoregressive decoding better suited?}
    \item \textbf{RQ2: Is stroke-5 the best initial representation?}
    \item \textbf{RQ3: How does the sketch order matter?}
\end{itemize}
Our findings across three tasks show that representing sketches as sequences of pen actions, as employed in Stroke-5, generally outperforms alternative approaches, with absolute pen coordinates proving more effective than relative ones in most cases. Furthermore, the importance of sketch ordering for the quality of the representations depends on task, order considered, and coordinate normalization. Finally, non-autoregressive decoding shows a superior reconstruction fidelity. Source code is available at \url{github.com/MarceloMoraesJr/Sketch-Temporality}.

\section{Preliminaries}
\label{sec:preliminaries}

\subsection{Sketch Definition}
\label{ss:sketch}
Sketch is a polyline comprised of multiple strokes with independent points. Usually, it is represented as a stroke-5 format \cite{SketchRNN, sketchformer}, defined as a sequence of $N$ two-dimensional coordinates $(x_i, y_i) \in \mathbb{R}^2$ and a three-binary pen state $\mathbf{p}_i \in \left\{ 0,1\right\}^3$, indicating whether a stroke is drawn, the end of a stroke, and the end of the sketch. Formally defined as
\begin{equation}
\left(x_0, y_0, \mathbf{p}_0\right),\; \left(\Delta{x_1}, \Delta{y_1}, \mathbf{p}_1\right),\; \dots,\; \left(\Delta{x_{N-1}},\; \Delta{y_{N-1}, \mathbf{p}_{N-1}}\right)
\end{equation}
where $(x_0, y_0) \in \mathbb{R}^2$ is the origin's absolute position, and $(\Delta{x_i},  \Delta{y_i}) = (x_{i} - x_{i-1}, y_{i} - y_{i-1})$ is the relative offset to the previous point. 


\subsection{Background}
\textbf{Sketch Representation Learning.} \citet{SketchINR} classifies approaches between parametric and non-parametric based on how sketches are modeled. Among non-parametric methods, SketchRNN \cite{SketchRNN} was the first to operate directly on vector sketches while preserving temporal order, later followed by SketchFormer \cite{sketchformer} and its hierarchical extension SketchFormer++ \cite{Sketchformer++}. Other non-parametric models such as SketchGNN and SketchSegFormer discard temporal information and rely solely on spatial cues. Parametric methods instead fit Bézier or spline curves to represent sketches while retaining stroke order \cite{BezierSketch, SketchCloud}. Most recently, SketchINR \cite{SketchINR} replaces discrete representations with implicit neural representations using a temporal parameterization.

\textbf{Relative or Absolute Coordinates?} Relative coordinates have been extensively used on different domains. For point cloud, allowing aggregating neighbor points in its local coordinate system \cite{PointNet++, PointTransformer, PointTransformerV2} or reframing patched points with respect to its centroid coordinates \cite{PointMAE}. Similarly for vision, using a local coordinate system for patches within the same window \cite{SwinTransformer}. Finally in text and graph domain, encoding relative distance between tokens and nodes \cite{RelativePosTransformer, Graphormer}. All in all, the relative coordinate system plays an important role towards introducing translation invariance of representations to other domains as done with CNNs for image pixels \cite{GDL, GSA, E(N)-GNN}. However, in sketch literature, there is a lack of consensus for what extent relative coordinates are better suited. Sequence-based methods stick to relative coordinates for tasks like sketch classification, segmentation, reconstruction and generation \cite{SketchRNN, sketchformer, Sketchformer++, Sketch-segformer, SketchKnitter}. Conversely, spatial-based methods advocate absolute coordinates for sketch image to vector \cite{bhunia2021vectorization} and capturing structural information for sketch segmentation \cite{SketchGNN, Sketch-segformer}.

\textbf{(Non-)Autoregressive Decoders.} Autoregressive (AR) decoding is often applied for sequence-based methods, based on the relevance of causal drawing order.
However, it inherits some limitations such as slow inference \cite{NAT}, noise propagation where prediction errors sum up throughout the sequence while predicting \cite{ErrorPropagation}, and exposure bias phenomenon due to the shift between training on teacher forcing and inferring feeding the own model's outputs \cite{ExposureBias1, ExposureBias2}. Specifically on the sketch completion task, AR decoders can only predict conditioned on the past coordinates \cite{SketchINR}. As a faster alternative, and discarding the relevance of causal decoding, non-autoregressive (NAR) decoding is employed in the literature \cite{NAT, MAE, PointMAE} and even for sketch domain \cite{SketchINR, StrokeCloud}. Thus, it casts a doubt on the true relevance of causal drawing order for reconstructing or generating sketches.

\textbf{Does the Sketch Order Really Matters?} Sequence-based approaches rely on the assumption that the temporal order of strokes carries meaningful information. SketchINR, for instance, explicitly incorporates stroke temporality, following evidence from \citet{TUBerlin} that certain sketch categories exhibit consistent stroke order, as well as insights from SketchRNN. SketchKnitter \cite{SketchKnitter} further reports that shuffling stroke order or intra-stroke points order during diffusion degrades the quality of generated sketches. Conversely, SketchFormer finds that permuting stroke order yields comparable performance for sketch classification. Moreover, spatial-based segmentation methods such as SketchGNN and SketchSegFormer disregard sketch order entirely. These contrasting observations highlight an open issue regarding which forms of sketch ordering matter and for which tasks.

\section{Methodology}
\label{sec:methodology}

We propose controlled modifications of a base architecture, evaluated on sketch tasks, to quantitatively assess representation quality.
\subsection{Architecture}

\textbf{Backbone.} We used the original Sketchformer autoencoder and encoder-only architectures for continuous input with some minor changes within the transformers blocks following ViTs \cite{ViT}, such as Pre-Layer Normalization and GELU activation for feed-forward network. The major changes rely on the decoder block, 
where we perform an ablation removing each component of the original transformer decoder: causal masking, making it NAR, cross-attention (CA) inspired by the AR decoder employed in Sketchformer++, and finally the self-attention (SA), leaving only the feed-forward network block, similar to the SketchINR decoder. For models without cross-attention, we simply sum the sketch-pooled representation from the encoder to the decoder's input position embeddings and tokens. 
It is noteworthy to mention that in order to make it fair, for feed-forward network only variants, we replaced the self-attention block with an extra feed-forward network block, given the parameter budget. 

\textbf{Sketch Tasks.} We trained for classification, segmentation, and reconstruction tasks. For the classification and segmentation, encoder-only models were trained via cross-entropy loss, with a linear classifier head applied on pooled representation for classification, and on point representations for segmentation. On the other hand, for the reconstruction task, we trained the whole model using only the L2 loss on the predicted coordinates. We used pen state (or other positional encodings) as known upfront to the decoder, following \citet{SketchINR,Sketchformer++} while focusing only on the reconstruction quality of coordinates.

\textbf{Input/Output Coordinates.} We investigated the use of both absolute and relative coordinates. Particularly for sketch reconstruction, there are model variants with different normalizations for outputs. Therefore, a model can reconstruct absolute coordinate points from relative coordinates or vice versa. 


\textbf{Sketch Positional Encoding.} We included the original three-way pen-state and the sinusoidal position embedding \cite{Transformers} as standard methods and studied new ones. The first is a local variation of the global sketch position for intra-stroke positions, namely, stroke position. Finally, as a substitute for the sequence pen-state, we also proposed a non-temporal based stroke embedding with a learnable matrix. All those positional embeddings are absolute in the sense that they are added to point coordinates at input level, compared to relative approaches, where they add a bias to the attention calculation \cite{PE_Extrapolation}. 

\textbf{Token Embedding.} The decoder of autoregressive methods has a token embedding block responsible for dealing with special tokens, e.g. \textit{start-of-sequence}, since input needs to be right-shifted. In non-autoregressive decoders, we have learnable tokens for every point to be reconstructed. That token is later enhanced with position information within the sketch.

\begin{figure}[tb]
    \centering
    \includegraphics[width=\columnwidth]{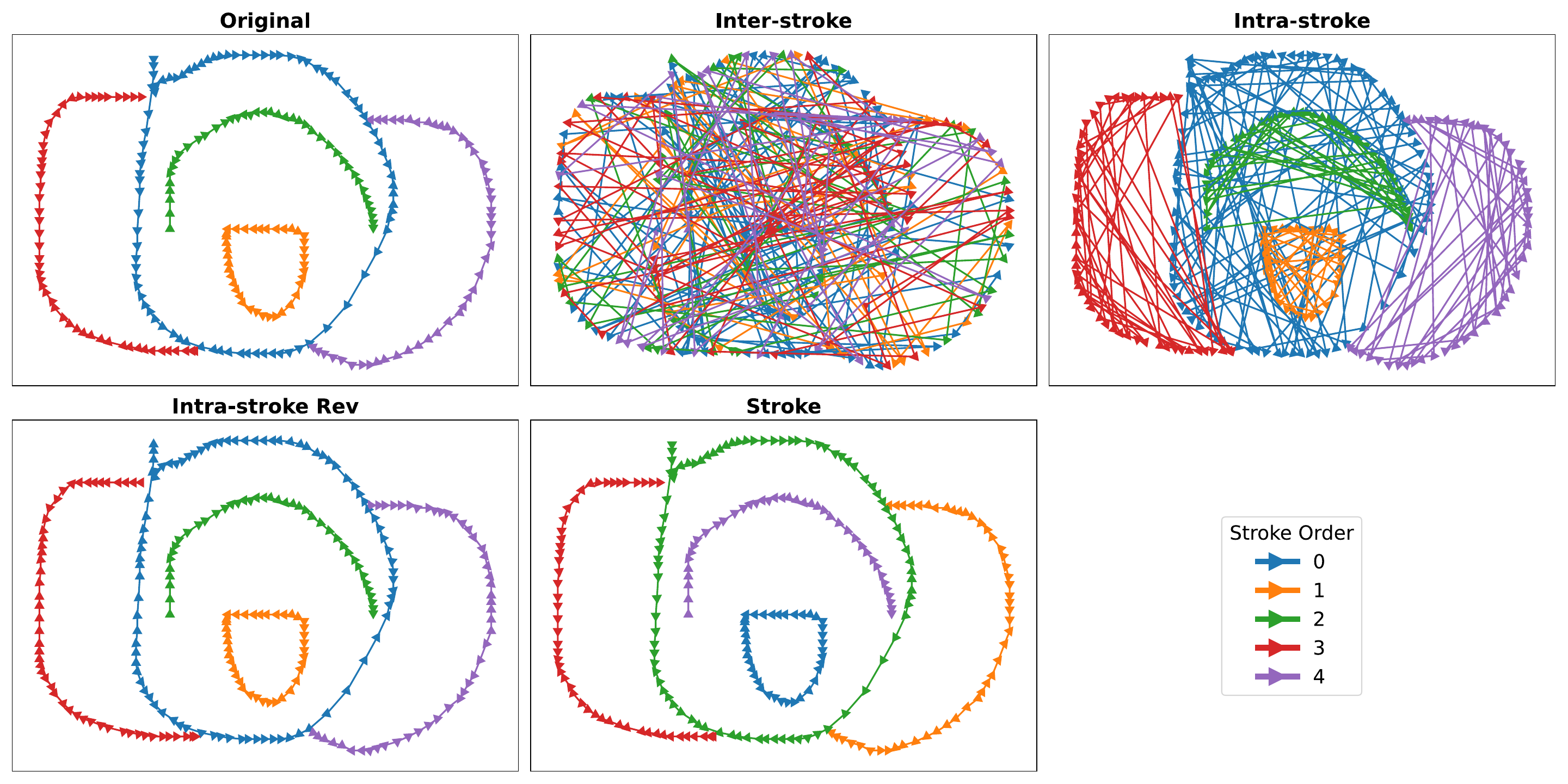}
    \caption{Proposed sketch order permutations. The triangle's position denotes the point's absolute coordinates, and it points to the next point following the intra-stroke order.}
    \label{fig:perturbations}
\end{figure}

\textbf{Sketch Permutation.} Variants of permutations were explored to investigate the quality of learned representations. First, an inter-stroke permutation as done by \citet{SketchKnitter}, which moves points from one to other strokes, and intra-stroke permutation of points. Also, intra-stroke reversion of the stroke drawing order. Finally, as in \citet{sketchformer, SketchKnitter}, shuffling the stroke order. All points and strokes are permutated with the exception of intra-stroke reversion, where half of the strokes are reversed at random to avoid the model to simply try to learn the reverse transformation. Figure \ref{fig:perturbations} presents all permutations.




\subsection{Evaluation} 

\textbf{Datasets.} We used the QuickDraw dataset \cite{SketchRNN} for sketch classification and reconstruction, which comprises 345 classes. Each class has 70k training examples, 2.5k for validation, and 2.5k for testing. We sampled 7k training examples for each class, approximately the same quantity used by \cite{sketchformer}. For sketch segmentation, we used the SPG dataset \cite{SPGDataset}, which is built on samples of QuickDraw. It comprises 25 categories with 800 sketches each, and a total of 109 stroke classes. Since the dataset is smaller, we build five random splits using the original ratio of 650 training instances, 50 validation, and 100 for testing for each category.  

\textbf{Metrics.} For sketch classification, we report test accuracy (Acc), while for reconstruction and segmentation, aiming for a test point-wise metric, we report, respectively, point mean squared error (MSE-point) and point accuracy (P-metric) \cite{SPGDataset}. For QuickDraw related tasks, we report the mean and standard deviation on the original test set of three random seeds. For segmentation on SPG, we report the mean and standard deviation on the five testing splits and five random seeds, one random seed for each split.

\textbf{Coordinate Normalization.} Regarding the relative coordinates, we follow the same implementation of \citet{SketchRNN, sketchformer}, which consists of min-max normalization by the maximum absolute coordinates difference. On the other hand, the unit circle normalization, well-known for point cloud processing, was applied over the absolute coordinates. Specifically for the sketch reconstruction task, models that output relative coordinates were later converted to absolute and normalized, once we evaluated them against an absolute coordinate ground truth.

\section{Experiments and Discussion}
\label{sec:experiments}


The experiments related to each research question are described in the following subsections.

\subsection{RQ1: Is autoregressive decoding better suited?}

We evaluated the impacts of removing different components from the Transformer decoder architecture and normalization variations in both input and output on the sketch reconstruction task via an autoencoder. Furthermore, we evaluated the knowledge transfer from pre-trained encoders in reconstruction to the classification task through a light finetuning of just one epoch on the same training set. Regarding positional embedding, the sketch position and pen-state patterns were employed. Table \ref{tab:decoder_type_recons} presents the results of all the variations investigated.

\begin{table*}
\renewcommand{\arraystretch}{1.3}
\centering
\setlength{\tabcolsep}{3pt}
\caption{Test pointwise MSE on sketch reconstruction and classification accuracy of fine-tuned encoders of decoder architectures and coordinate normalizations for input and output. Symbol ($\boldsymbol{-}$) denotes the removal of a component in the decoder block. References are encoders trained from scratch under the same conditions. The best results for each coordinate normalization input are highlighted in bold.}
\vspace{1em}
\label{tab:decoder_type_recons}
{\small    
  \begin{tabular}{clcccc}
    \hline
    & & \multicolumn{2}{c}{Relative Input} & \multicolumn{2}{c}{Absolute Input} \\
    Task & Decoder Arch. &  Relative Output  & Absolute Output & Relative Output & Absolute Output \\
    \hline
    \multirow{5}{*}{\makecell{Reconstruction\\(MSE-Point x$10^{-2}$)}} & AR & 63.359$\pm$05.546 & 51.465$\pm$06.991 & 81.222$\pm$10.396 & 25.523$\pm$02.941\\
    & \hspace{1.0em}$\boldsymbol{-}$ CA & 58.503$\pm$05.394 & 40.762$\pm$03.956 & 75.842$\pm$04.528 & 21.146$\pm$01.337 \\
    & NAR & 04.460$\pm$01.539 & 00.452$\pm$00.012 & 02.875$\pm$00.178 & 00.306$\pm$00.018\\
    & \hspace{1.0em}$\boldsymbol{-}$ CA & 04.182$\pm$00.726 & \textbf{00.376$\pm$00.008} & 02.737$\pm$00.863	& \textbf{00.233$\pm$00.008}\\
    & \hspace{1.0em}$\boldsymbol{-}$ CA $\boldsymbol{-}$ SA & 10.423$\pm$02.512 & 00.462$\pm$00.004 & 04.404$\pm$00.213 & 00.261$\pm$00.005 \\
    \hline
    \multirow{6}{*}{\makecell{Fine-tuning\\(Acc)}} & AR & 66.49$\pm$03.67 & \textbf{71.70$\pm$00.68} & 67.28$\pm$02.76 & \textbf{71.98$\pm$00.17} \\
    & \hspace{1.0em}$\boldsymbol{-}$ CA & 64.64$\pm$01.27 & 70.94$\pm$01.59 & 66.15$\pm$03.48 & 71.38$\pm$00.87 \\
    & NAR & 66.73$\pm$00.53 & 69.24$\pm$00.49 & 69.74$\pm$00.99 & 70.50$\pm$00.34 \\
    & \hspace{1.0em}$\boldsymbol{-}$ CA & 64.44$\pm$00.63 & 66.86$\pm$00.25 & 68.86$\pm$01.28 & 69.45$\pm$00.48 \\
    & \hspace{1.0em}$\boldsymbol{-}$ CA $\boldsymbol{-}$ SA & 52.41$\pm$00.35 & 68.23$\pm$00.28 & 67.96$\pm$00.93 & 68.73$\pm$01.06\\
    & Reference & \multicolumn{2}{c}{64.36$\pm$00.70} & \multicolumn{2}{c}{64.48$\pm$00.62}\\
    \hline
  \end{tabular}
}    
  
\end{table*}

\begin{figure}
    \centering
    \includegraphics[width=\columnwidth]{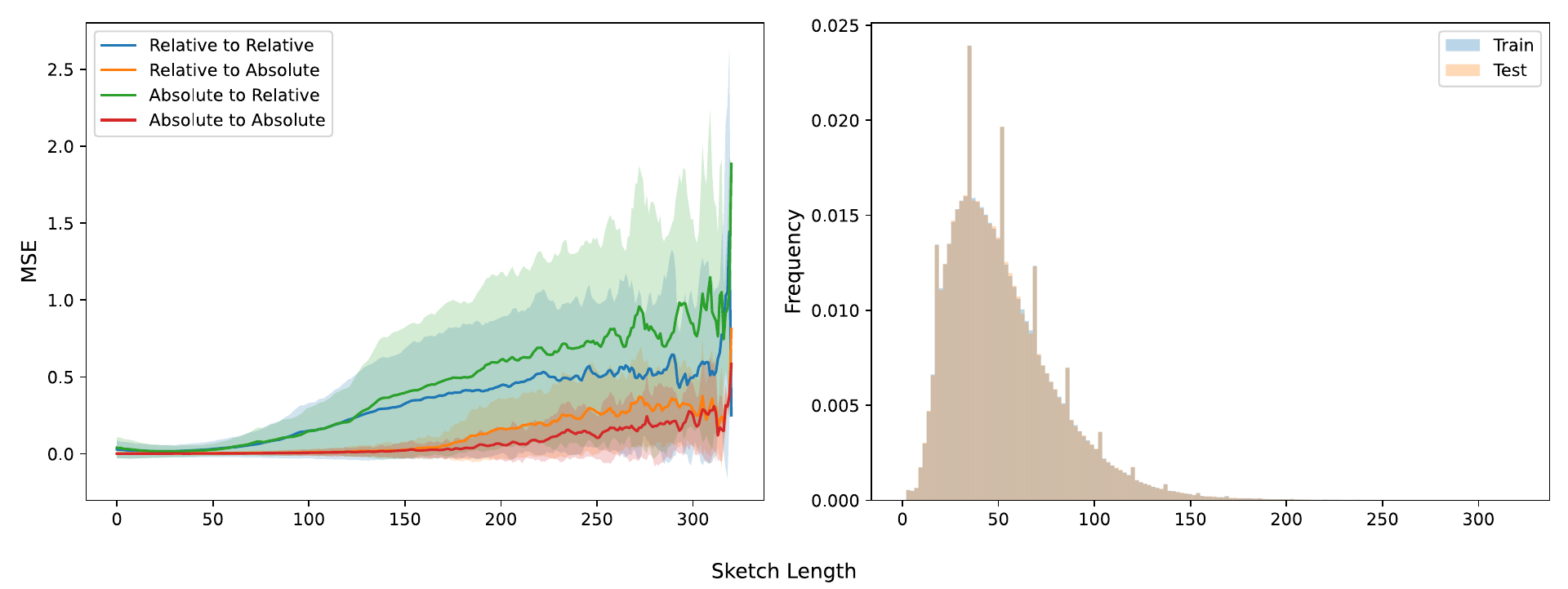}
    \caption{(Left) Mean MSE and standard deviation (lower is better) of each sketch position on test set. (Right) Train and test set distributions of sketch lengths. Distributions are similarly right-skewed due to rare long sequences.}
    \label{fig:prop_erros}
\end{figure}

Overall, using absolute coordinates as the architecture's output resulted in better reconstructions, especially when combined with inputs of the same normalization. This highlights the inherent problem with relative coordinates: error propagation due to the cumulative sum operation when converting the predicted sketch to the absolute one. Figure \ref{fig:prop_erros} supports this argument, where relative coordinates register a higher average error and greater variability for the last points of the sketch. This effect is also exacerbated by the smaller number of long sketches. To a lesser extent, the same effect is observed for absolute coordinates and is due to the use of sketch position as an absolute sinusoidal, which consequently makes it difficult for the model to extrapolate to longer sketches than those frequently seen during training \cite{PE_Extrapolation}.

Regarding decoders, autoregressive decoders demonstrated difficulty in generalization due to the effect of exposure bias. This is heightened by the continuous nature of the coordinates, which is supported by the lower metrics obtained compared to the discretized versions in the original Sketchformer study \cite{sketchformer}. Furthermore, the use of cross-attention proved unnecessary once it attends only to the pooled representation. It was surpassed by a simple strategy of adding the representation to the input tokens and positional embeddings, achieving the best reconstruction results. Finally, for non-autoregressive decoders, the use of a feed-forward only decoder is outperformed by its combination with self-attention. In particular, its use may still be a viable alternative for outputting absolute coordinates due to its linear cost, as opposed to the quadratic cost of self-attention. However, for relative coordinates, it performed with an MSE approximately twice as high as other non-autoregressive decoders. This result suggests the need for a mechanism for reconstructing relative relationships present in this type of normalization.

In knowledge transfer, absolute coordinates maintained their superiority over relative ones. However, in contrast to the reconstruction results, encoders pretrained with decoders that achieved better reconstruction quality led to lower adaptivity. This result reinforces the need for strategies that regularize the latent space of autoencoders, especially the family of denoising autoencoders \cite{DAE}, commonly used for self-supervised representation learning in other domains such as text \cite{BERT, GPT} and images \cite{MAE}.
In Table \ref{tab:denoising}, we convert autoencoders from absolute coordinates to absolute for denoising, with the addition of Gaussian noise and random dropout rate at the input points coordinates. Consequently, an increase in the accuracy of the downstream task was recorded, especially for decoders with greater reconstruction quality, showing this to be a promising direction in sketch self-supervised representation learning.

\begin{table}
\centering
\caption{Downstream classification accuracies of pretrained encoders in denoising autoencoder compared to classical of different decoder architectures. All variants use absolute to absolute coordinates. Symbol ($\boldsymbol{-}$) denotes the removal of a component in decoder block.}
  \label{tab:denoising}
  \vspace{1em}
\renewcommand{\arraystretch}{1.3}
\setlength{\tabcolsep}{3pt}
  {\small
  \begin{tabular}{lcc}
    \hline
    & \multicolumn{2}{c}{Downstream Acc}\\
    Decoder Arch. & Classical & Denoising \\
    \hline
    AR & 71.98$\pm$00.17 & 70.07$\pm$00.99\\
    \hspace{1em} $\boldsymbol{-}$ CA & 71.38$\pm$00.87 & 71.51$\pm$00.08\\
    NAR & 70.50$\pm$00.34 & 70.39$\pm$00.17 \\
    \hspace{1em} $\boldsymbol{-}$ CA & 69.45$\pm$00.48 & 71.20$\pm$00.72\\
    \hspace{1em} $\boldsymbol{-}$ CA $\boldsymbol{-}$ SA & 68.73$\pm$01.06 & 70.26$\pm$00.42\\
    \hline
  \end{tabular}
}
\end{table}

\subsection{RQ2: Is stroke-5 the best initial representation?}

To evaluate the impact of different positional encodings and coordinate normalization on sketch representation learning, we verified in Table \ref{tab:pe} the variations discussed in the previous section across the three tasks. Specifically for the sketch reconstruction task, we used the non-autoregressive decoder without cross-attention with absolute coordinate output due to the better results obtained in the previous experiment.

\begin{table*}[h]
\renewcommand{\arraystretch}{1.3}
\centering
\setlength{\tabcolsep}{3pt}
\caption{Sketch positional encodings and coordinate normalizations performance on multiple sketch tasks. Each task has its own metric in parenthesis. Symbol ($\boldsymbol{+}$) denotes combination of positional encodings. Best positional informations for each coordinate normalization and task is highlightned in bold.}
\label{tab:pe}
\vspace{1em}
{\small
\begin{tabular}{lcccccc}
\hline
& \multicolumn{2}{c}{Classification (Acc)} & \multicolumn{2}{c}{Segmentation (P-metric)} & \multicolumn{2}{c}{Reconstruction (MSE-point x $10^{-2}$)}\\
Positional Encoding & Relative & Absolute  & Relative & Absolute  & Relative & Absolute  \\
\hline
None & 26.96$\pm$12.97 & 56.02$\pm$07.25 & 18.61$\pm$00.42 & 64.53$\pm$00.76 & 34.609$\pm$00.000 & 34.609$\pm$00.000\\
Stroke Pos. & 64.89$\pm$00.50 & 75.78$\pm$00.05 & 40.60$\pm$05.05 & 78.45$\pm$02.34 & 20.169$\pm$00.026 & 18.471$\pm$00.011 \\
Sketch Pos. & 80.53$\pm$00.17 & 78.93$\pm$00.19 & 40.69$\pm$00.72 & 80.43$\pm$02.75 & 00.442$\pm$00.020 & 00.303$\pm$00.008\\
Stroke Emb. & 63.49$\pm$02.82 & 75.75$\pm$00.21 & 47.72$\pm$02.62 & 87.03$\pm$01.43 & 22.702$\pm$00.034 & 20.123$\pm$00.000\\
Stroke Emb. $\boldsymbol{+}$ Stroke Pos. & 80.71$\pm$00.06 & 80.72$\pm$00.16 & \textbf{53.84$\pm$00.54} & 86.48$\pm$01.75 & 00.646$\pm$00.015 & 00.377$\pm$00.009\\
Stroke Emb. $\boldsymbol{+}$ Sketch Pos. & 80.78$\pm$00.09 & 80.52$\pm$00.21 & 49.20$\pm$01.39 & 84.61$\pm$01.30 & 00.670$\pm$00.038 & 00.316$\pm$00.006\\
Pen-State & 46.40$\pm$00.81 & 70.66$\pm$00.79 & 21.78$\pm$01.10 & 68.17$\pm$01.50 & 33.583$\pm$00.004 & 33.047$\pm$00.001\\
Pen-State $\boldsymbol{+}$ Stroke Pos. & 69.14$\pm$00.73 & 77.46$\pm$00.24 & 41.72$\pm$02.14 & 82.71$\pm$00.47 & 18.447$\pm$00.030 & 16.299$\pm$00.008\\
Pen-State $\boldsymbol{+}$ Sketch Pos. & \textbf{81.49$\pm$00.25} & \textbf{81.00$\pm$00.06} & 48.05$\pm$01.23 & \textbf{87.19$\pm$01.19} & \textbf{00.376$\pm$00.008} & \textbf{00.233$\pm$00.008} \\
\hline
\end{tabular}
}
\end{table*}

In classification and reconstruction tasks, absolute normalization is superior to relative normalization when using variations that do not uniquely identify all points to be reconstructed, such as none, stroke position, stroke embedding, pen-state, and stroke position. This is explained by the fact that absolute coordinates uniquely identify a single point, while the same relative coordinate can identify multiple points, and the sequence order defines their meaning. However, in sketch reconstruction, due to the use of a NAR decoder, independent of normalization, unique positioning information is required to condition the decoder to reconstruct the exact point. Therefore, especially in this task, non-unique variations led to a greater increase in the metric.

Particularly in the segmentation task, absolute normalization is superior in all position embedding scenarios, achieving a result approximately 1.6 times higher when compared to its best variants. These results, combined with the small scale of the dataset, suggest that absolute coordinates offer a bias in the separation of stroke classes. This same superiority of absolute coordinates as input over relative coordinates was observed in the reconstruction task independent of positional encoding, which is supported by the previous experiment.

Among the positional encoding variations tested, the original pen-state with sketch position stands out, achieving better metrics in all three tasks and in both normalizations. That reflects the importance of treating the sketch as a global sequence, as methods in the literature usually address, and the pen-state ability to be invariant to number and order of strokes. In this best variation, it is worth noting that relative coordinates obtained better results than absolute coordinates in recognizing sketch classes, which shows the translation invariance of relative offsets. Therefore is a promising direction of investigating true translation equivariance and the combination of both, as seen in point clouds models \cite{PointNet, PointNet++, PointTransformer, PointTransformerV2}.

\subsection{RQ3: Does the sketch order matter?}

Given the best decoder for reconstruction and positional encodings for each normalization and task, we evaluated the importance of sketch drawing order in learning sketch representation through the proposed order permutations. Table \ref{tab:order_perturbation} summarizes the results of this study.

\begin{table*}
\renewcommand{\arraystretch}{1.3}
\centering
\setlength{\tabcolsep}{3pt}
\caption{Sketch order permutations impact on each task. Each task has its own metric in parenthesis. Symbol ($\boldsymbol{+}$) denotes combination of order permutations.}
\label{tab:order_perturbation}
\vspace{1em}
{\small
\begin{tabular}{lcccccc}
\hline
& \multicolumn{2}{c}{Classification (Acc)} & \multicolumn{2}{c}{Segmentation (P-metric)} & \multicolumn{2}{c}{Reconstruction (MSE-point x $10^{-2}$)}\\
Order Permutation & Relative & Absolute  & Relative & Absolute  & Relative & Absolute  \\
\hline
None & 81.49$\pm$00.25 & 81.00$\pm$00.06 & 53.84$\pm$00.54 & 87.19$\pm$01.19 & 00.376$\pm$00.008 & 00.233$\pm$00.008\\
Inter-Stroke & 57.36$\pm$00.45 & 69.19$\pm$00.09 & 32.49$\pm$00.75 & 65.36$\pm$00.36 & 04.902$\pm$00.398 & 03.699$\pm$00.286\\
Intra-Stroke & 73.26$\pm$00.22 & 77.31$\pm$00.10 & 67.67$\pm$01.30 & 86.20$\pm$02.45 & 02.030$\pm$00.092 & 01.595$\pm$00.092\\
Intra-Stroke Rev. & 80.78$\pm$00.19 & 80.42$\pm$00.23 & 46.76$\pm$01.83 & 85.86$\pm$01.65 & 00.393$\pm$00.015 & 00.229$\pm$00.008\\
Stroke & 80.11$\pm$00.27 & 79.98$\pm$00.14 & 39.72$\pm$03.49 & 82.81$\pm$01.56 & 00.404$\pm$00.017 & 00.228$\pm$00.006\\
Stroke $\boldsymbol{+}$ Intra-Stroke & 71.80$\pm$00.08 & 76.37$\pm$00.12 & 63.06$\pm$01.58 & 86.31$\pm$01.52 & 02.491$\pm$00.097 & 01.857$\pm$00.084\\
Stroke  $\boldsymbol{+}$ Intra-Stroke Rev. & 79.82$\pm$00.26 & 79.51$\pm$00.21 & 33.15$\pm$04.05 & 83.18$\pm$01.20 & 00.398$\pm$00.016 & 00.232$\pm$00.009\\

\hline
\end{tabular}
}
\end{table*}

Permutation of points between strokes had the greatest impact, as it introduces the largest amount of noise, completely altering the stroke structures. Even so, the models performed better than the variation without the addition of positional encoding in the previous experiment. Meanwhile, permuting the order of points within strokes hindered classification accuracies and reconstruction L2, while having no significant impact on segmentation. Its use also promoted an increase in performance in segmentation with relative coordinates, which can be seen as a data augmentation in this small-scale dataset. Reversing the order of points within a stroke did not lead to a significant drop in performance for all tasks, except for relative coordinates in segmentation and reconstruction. However, rearranging the order of strokes in the sketch led to a reduction in metrics and, combined with other permutations, exacerbated this effect. The exceptions were for segmentation, where the combination with intra-stroke augmentation property reversed the effect, and for absolute coordinates on reconstruction task, where it did not impact the model's reconstruction quality.

The results suggest that the usefulness of temporality depends on the task and the normalization. However, it is still unclear how much the results are due to the sinusoidal-based position encoding, as it is non-equivariant to translation and permutation \cite{GSA}.


\section{Conclusion}
\label{sec:conclusion}

Overall, the relationship between point order within strokes and the use of positional encodings for pen state and sketch position, both inspired by human motor patterns, are important factors for representation quality. Yet, our results show that relative coordinates are consistently outperformed by absolute ones across most tasks. Moreover, although temporal structure matters, non-autoregressive decoders are superior for reconstruction and downstream evaluation. Researchers can apply these findings by favoring absolute coordinate encodings, reconsidering the need for strict temporal modeling, and employing non-autoregressive architectures. We believe these insights may also support advances in sketch generation.




Future work may include investigating methods to reduce the exposure bias inherent to autoregressive decoders, exploring alternatives to sinusoidal absolute positional encodings, and examining coordinate normalization through the lens of geometric equivariance. Finally, hybrid approaches that integrate the strengths of multiple encoding strategies remain an open and promising direction for further research.

\section*{Acknowledgements}
This study was financed in part by the Coordenação de Aperfeiçoamento de Pessoal de Nível Superior - Brasil (CAPES) - Finance Code 001. We thank Prof. Leo Sampaio Ferraz Ribeiro for useful early discussions.

\bibliographystyle{elsarticle-num-names} 
\bibliography{references}

@inproceedings{SketchRNN,
    title={A Neural Representation of Sketch Drawings},
    author={David Ha and Douglas Eck},
    booktitle={ICLR},
    year={2018},
    urlX={https://openreview.net/forum?id=Hy6GHpkCW},
}

@inproceedings{Sketchformer,
  title={Sketchformer: Transformer-based representation for sketched structure},
  author={Ribeiro, Leo Sampaio Ferraz and Bui, Tu and Collomosse, John and Ponti, Moacir},
  booktitle={CVPR},
  pagesX={14153--14162},
  year={2020}
}

@inproceedings{Sketchformer++,
    author = {Xu, Pengfei and Ruan, Banhuai and Zheng, Youyi and Huang, Hui},
    title = {Sketchformer++: A Hierarchical Transformer Architecture for Vector Sketch Representation},
    year = {2024},
    isbnX = {978-981-97-2094-1},
    publisherX = {Springer-Verlag},
    addressX = {Berlin, Heidelberg},
    urlX = {https://doi.org/10.1007/978-981-97-2095-8_2},
    doiX = {10.1007/978-981-97-2095-8_2},
    abstractX = {With the rising ubiquity of digital touch devices and sketch-based interfaces, freehand sketching has become an essential mode of visual communication. Nevertheless, interpreting these often ambiguous and sparse sketches poses challenges for computers. This paper presents Sketchformer++, a hierarchical transformer architecture for the neural representation of vector sketches. It treats a vector sketch as a three-level structure, i.e., sketch level, stroke level, and segment level. Three self-attention modules are adopted in the network architecture, corresponding to the sketch hierarchy. The semantics of sketches are aggregated from local to global, resulting in neural representations of sketches. Extensive experiments show that Sketchformer++ exhibits superior performance in various downstream tasks, including sketch reconstruction, sketch recognition, and sketch semantic segmentation, demonstrating its robustness and effectiveness in sketch representation.},
    booktitle = {ICCVM},
    pagesX = {24–41},
    numpagesX = {18},
    keywordsX = {Vector sketch, Transformer, Hierarchy, Neural representation, Sketch recognition, Sketch semantic segmentation},
    locationX = {Wellington, New Zealand}
}

@INPROCEEDINGS{SketchINR,
  author={Bandyopadhyay, Hmrishav and Bhunia, Ayan Kumar and Chowdhury, Pinaki Nath and Sain, Aneeshan and Xiang, Tao and Hospedales, Timothy and Song, Yi-Zhe},
  booktitle={CVPR}, 
  title={SketchINR: A First Look into Sketches as Implicit Neural Representations}, 
  year={2024},
  volume={},
  number={},
  pagesX={12565-12574},
  keywordsX={Visualization;Interpolation;Computer vision;Shape;Computational modeling;Data compression;Vectors},
  doiX={10.1109/CVPR52733.2024.01194}
}

@inproceedings{PointNet++,
    author = {Qi, Charles R. and Yi, Li and Su, Hao and Guibas, Leonidas J.},
    title = {PointNet++: deep hierarchical feature learning on point sets in a metric space},
    year = {2017},
    isbnX = {9781510860964},
    publisherX = {Curran Associates Inc.},
    addressX = {Red Hook, NY, USA},
    abstractX = {Few prior works study deep learning on point sets. PointNet [20] is a pioneer in this direction. However, by design PointNet does not capture local structures induced by the metric space points live in, limiting its ability to recognize fine-grained patterns and generalizability to complex scenes. In this work, we introduce a hierarchical neural network that applies PointNet recursively on a nested partitioning of the input point set. By exploiting metric space distances, our network is able to learn local features with increasing contextual scales. With further observation that point sets are usually sampled with varying densities, which results in greatly decreased performance for networks trained on uniform densities, we propose novel set learning layers to adaptively combine features from multiple scales. Experiments show that our network called PointNet++ is able to learn deep point set features efficiently and robustly. In particular, results significantly better than state-of-the-art have been obtained on challenging benchmarks of 3D point clouds.},
    booktitle = {NeurIPS},
    pagesX = {5105–5114},
    numpagesX = {10},
    locationX = {Long Beach, California, USA},
    seriesX = {NIPS'17}
}

@InProceedings{PointTransformer,
    author    = {Zhao, Hengshuang and Jiang, Li and Jia, Jiaya and Torr, Philip H.S. and Koltun, Vladlen},
    title     = {Point Transformer},
    booktitle = {ICCV},
    monthX     = {October},
    year      = {2021},
    pagesX     = {16259-16268}
}

@inproceedings{PointTransformerV2,
 author = {Wu, Xiaoyang and Lao, Yixing and Jiang, Li and Liu, Xihui and Zhao, Hengshuang},
 booktitle = {NeurIPS},
 editorX = {S. Koyejo and S. Mohamed and A. Agarwal and D. Belgrave and K. Cho and A. Oh},
 pagesX = {33330--33342},
 publisherX = {Curran Associates, Inc.},
 title = {Point Transformer V2: Grouped Vector Attention and Partition-based Pooling},
 urlX = {https://proceedings.neurips.cc/paper_files/paper/2022/file/d78ece6613953f46501b958b7bb4582f-Paper-Conference.pdf},
 volumeX = {35},
 year = {2022}
}

@InProceedings{PointMAE,
    author="Pang, Yatian
    and Wang, Wenxiao
    and Tay, Francis E. H.
    and Liu, Wei
    and Tian, Yonghong
    and Yuan, Li",
    editorX="Avidan, Shai
    and Brostow, Gabriel
    and Ciss{\'e}, Moustapha
    and Farinella, Giovanni Maria
    and Hassner, Tal",
    title="Masked Autoencoders for Point Cloud Self-supervised Learning",
    booktitle="ECCV",
    year="2022",
    publisherX="Springer Nature Switzerland",
    addressX="Cham",
    pagesX="604--621",
    abstractX="As a promising scheme of self-supervised learning, masked autoencoding has significantly advanced natural language processing and computer vision. Inspired by this, we propose a neat scheme of masked autoencoders for point cloud self-supervised learning, addressXing the challenges posed by point cloud's properties, including leakage of locationX information and uneven information density. Concretely, we divide the input point cloud into irregular point patches and randomly mask them at a high ratio. Then, a standard Transformer based autoencoder, with an asymmetric design and a shifting mask tokens operation, learns high-level latent features from unmasked point patches, aiming to reconstruct the masked point patches. Extensive experiments show that our approach is efficient during pre-training and generalizes well on various downstream tasks. The pre-trained models achieve 85.18{\%} accuracy on ScanObjectNN and 94.04{\%} accuracy on ModelNet40, outperforming all the other self-supervised learning methods. We show with our scheme, a simple architecture entirely based on standard Transformers can surpass dedicated Transformer models from supervised learning. Our approach also advances state-of-the-art accuracies by 1.5{\%}--2.3{\%} in the few-shot classification. Furthermore, our work inspires the feasibility of applying unified architectures from languages and images to the point cloud. Codes are available at https://github.com/Pang-Yatian/Point-MAE.",
    isbnX="978-3-031-20086-1"
}

@InProceedings{SwinTransformer,
    author    = {Liu, Ze and Lin, Yutong and Cao, Yue and Hu, Han and Wei, Yixuan and Zhang, Zheng and Lin, Stephen and Guo, Baining},
    title     = {Swin Transformer: Hierarchical Vision Transformer Using Shifted Windows},
    booktitle = {ICCV},
    monthX     = {October},
    year      = {2021},
    pagesX     = {10012-10022}
}

@inproceedings{RelativePosTransformer,
    title = "Self-Attention with Relative Position Representations",
    author = "Shaw, Peter  and
      Uszkoreit, Jakob  and
      Vaswani, Ashish",
    editorX = "Walker, Marilyn  and
      Ji, Heng  and
      Stent, Amanda",
    booktitle = "NAACL",
    monthX = jun,
    year = "2018",
    addressX = "New Orleans, Louisiana",
    publisherX = "Association for Computational Linguistics",
    urlX = "https://aclanthology.org/N18-2074/",
    doiX = "10.18653/v1/N18-2074",
    pagesX = "464--468",
    abstractX = "Relying entirely on an attention mechanism, the Transformer introduced by Vaswani et al. (2017) achieves state-of-the-art results for machine translation. In contrast to recurrent and convolutional neural networks, it does not explicitly model relative or absolute position information in its structure. Instead, it requires adding representations of absolute positions to its inputs. In this work we present an alternative approach, extending the self-attention mechanism to efficiently consider representations of the relative positions, or distances between sequence elements. On the WMT 2014 English-to-German and English-to-French translation tasks, this approach yields improvements of 1.3 BLEU and 0.3 BLEU over absolute position representations, respectively. Notably, we observe that combining relative and absolute position representations yields no further improvement in translation quality. We describe an efficient implementation of our method and cast it as an instance of relation-aware self-attention mechanisms that can generalize to arbitrary graph-labeled inputs."
}

@inproceedings{Graphormer,
    author = {Ying, Chengxuan and Cai, Tianle and Luo, Shengjie and Zheng, Shuxin and Ke, Guolin and He, Di and Shen, Yanming and Liu, Tie-Yan},
    title = {Do transformers really perform bad for graph representation?},
    year = {2021},
    isbnX = {9781713845393},
    publisherX = {Curran Associates Inc.},
    addressX = {Red Hook, NY, USA},
    abstractX = {The Transformer architecture has become a dominant choice in many domains, such as natural language processing and computer vision. Yet, it has not achieved competitive performance on popular leaderboards of graph-level prediction compared to mainstream GNN variants. Therefore, it remains a mystery how Transformers could perform well for graph representation learning. In this paper, we solve this mystery by presenting Graphormer, which is built upon the standard Transformer architecture, and could attain excellent results on a broad range of graph representation learning tasks, especially on the recent OGB Large-Scale Challenge. Our key insight to utilizing Transformer in the graph is the necessity of effectively encoding the structural information of a graph into the model. To this end, we propose several simple yet effective structural encoding methods to help Graphormer better model graph-structured data. Besides, we mathematically characterize the expressive power of Graphormer and exhibit that with our ways of encoding the structural information of graphs, many popular GNN variants could be covered as the special cases of Graphormer.},
    booktitle = {NeurIPS},
    articlenoX = {2212},
    numpagesX = {12},
    seriesX = {NIPS '21}
}

@inproceedings{Transformers,
    author = {Vaswani, Ashish and Shazeer, Noam and Parmar, Niki and Uszkoreit, Jakob and Jones, Llion and Gomez, Aidan N. and Kaiser, \L{}ukasz and Polosukhin, Illia},
    title = {Attention is all you need},
    year = {2017},
    isbnX = {9781510860964},
    publisherX = {Curran Associates Inc.},
    addressX = {Red Hook, NY, USA},
    abstractX = {The dominant sequence transduction models are based on complex recurrent or convolutional neural networks that include an encoder and a decoder. The best performing models also connect the encoder and decoder through an attention mechanism. We propose a new simple network architecture, the Transformer, based solely on attention mechanisms, dispensing with recurrence and convolutions entirely. Experiments on two machine translation tasks show these models to be superior in quality while being more parallelizable and requiring significantly less time to train. Our model achieves 28.4 BLEU on the WMT 2014 English-to-German translation task, improving over the existing best results, including ensembles, by over 2 BLEU. On the WMT 2014 English-to-French translation task, our model establishes a new single-model state-of-the-art BLEU score of 41.0 after training for 3.5 days on eight GPUs, a small fraction of the training costs of the best models from the literature.},
    booktitle = {NeurIPS},
    pagesX = {6000–6010},
    numpagesX = {11},
    locationX = {Long Beach, California, USA},
    seriesX = {NIPS'17}
}

@InProceedings{bhunia2021vectorization,
    author    = {Bhunia, Ayan Kumar and Chowdhury, Pinaki Nath and Yang, Yongxin and Hospedales, Timothy M. and Xiang, Tao and Song, Yi-Zhe},
    title     = {Vectorization and Rasterization: Self-Supervised Learning for Sketch and Handwriting},
    booktitle = {CVPR},
    monthX     = {June},
    year      = {2021},
    pagesX    = {5672-5681}
}

@inproceedings{SketchKnitter,
    title={SketchKnitter: Vectorized Sketch Generation with Diffusion Models},
    author={Qiang Wang and Haoge Deng and Yonggang Qi and Da Li and Yi-Zhe Song},
    booktitle={ICLR},
    year={2023},
    urlX={https://openreview.net/forum?id=4eJ43EN2g6l}
}

@ARTICLE{Sketch-segformer,
  author={Zheng, Yixiao and Xie, Jiyang and Sain, Aneeshan and Song, Yi-Zhe and Ma, Zhanyu},
  journal={IEEE Trans. Image Process.}, 
  title={Sketch-Segformer: Transformer-Based Segmentation for Figurative and Creative Sketches}, 
  year={2023},
  volumeX={32},
  numberX={},
  pagesX={4595-4609},
  keywordsX={Semantics;Semantic segmentation;Transformers;Task analysis;Solid modeling;Feature extraction;Stroke (medical condition);Sketch semantic segmentation;2D structured point set representation;dual self-attention block;order embedding;sketch-specific transformer-based framework},
  doiX={10.1109/TIP.2023.3302521}
}

@ARTICLE{SPGDataset,
  author={Li, Ke and Pang, Kaiyue and Song, Yi-Zhe and Xiang, Tao and Hospedales, Timothy M. and Zhang, Honggang},
  journal={IEEE Trans. Image Process.}, 
  title={Toward Deep Universal Sketch Perceptual Grouper}, 
  year={2019},
  volumeX={28},
  numberX={7},
  pagesX={3219-3231},
  keywordsX={Semantics;Image segmentation;Task analysis;Visualization;Training;Data models;Analytical models;Sketch perceptual grouping;universal grouper;deep grouping model;dataset},
  doiX={10.1109/TIP.2019.2895155}
}

@inproceedings{NAT,
    title={Non-Autoregressive Neural Machine Translation},
    author={Jiatao Gu and James Bradbury and Caiming Xiong and Victor O.K. Li and Richard Socher},
    booktitle={ICLR},
    year={2018},
    urlX={https://openreview.net/forum?id=B1l8BtlCb},
}

@inproceedings{ExposureBias1,
    author = {Bengio, Samy and Vinyals, Oriol and Jaitly, Navdeep and Shazeer, Noam},
    title = {Scheduled sampling for sequence prediction with recurrent Neural networks},
    year = {2015},
    publisherX = {MIT Press},
    addressX = {Cambridge, MA, USA},
    abstractX = {Recurrent Neural Networks can be trained to produce sequences of tokens given some input, as exemplified by recent results in machine translation and image captioning. The current approach to training them consists of maximizing the likelihood of each token in the sequence given the current (recurrent) state and the previous token. At inference, the unknown previous token is then replaced by a token generated by the model itself. This discrepancy between training and inference can yield errors that can accumulate quickly along the generated sequence. We propose a curriculum learning strategy to gently change the training process from a fully guided scheme using the true previous token, towards a less guided scheme which mostly uses the generated token instead. Experiments on several sequence prediction tasks show that this approach yields significant improvements. Moreover, it was used succesfully in our winning entry to the MSCOCO image captioning challenge, 2015.},
    booktitle = {NeurIPS},
    pagesX = {1171–1179},
    numpagesX = {9},
    locationX = {Montreal, Canada},
    seriesX = {NIPS'15}
}

@inproceedings{ExposureBias2,
  author={Marc'Aurelio Ranzato and Sumit Chopra and Michael Auli and Wojciech Zaremba},
  title={Sequence Level Training with Recurrent Neural Networks},
  year={2016},
  cdateX={1451606400000},
  urlX={http://arxiv.org/abs/1511.06732},
  booktitle={ICLR}
}

@inproceedings{ErrorPropagation,
    title = "Beyond Error Propagation in Neural Machine Translation: Characteristics of Language Also Matter",
    author = "Wu, Lijun  and
      Tan, Xu  and
      He, Di  and
      Tian, Fei  and
      Qin, Tao  and
      Lai, Jianhuang  and
      Liu, Tie-Yan",
    editorX = "Riloff, Ellen  and
      Chiang, David  and
      Hockenmaier, Julia  and
      Tsujii, Jun{'}ichi",
    booktitle = "EMNLP",
    monthX = oct # "-" # nov,
    year = "2018",
    addressX = "Brussels, Belgium",
    publisherX = "Association for Computational Linguistics",
    urlX = "https://aclanthology.org/D18-1396/",
    doiX = "10.18653/v1/D18-1396",
    pagesX = "3602--3611",
    abstractX = "Neural machine translation usually adopts autoregressive models and suffers from exposure bias as well as the consequent error propagation problem. Many previous works have discussed the relationship between error propagation and the \textit{accuracy drop} (i.e., the left part of the translated sentence is often better than its right part in left-to-right decoding models) problem. In this paper, we conduct a series of analyses to deeply understand this problem and get several interesting findings. (1) The role of error propagation on accuracy drop is overstated in the literature, although it indeed contributes to the accuracy drop problem. (2) Characteristics of a language play a more important role in causing the accuracy drop: the left part of the translation result in a right-branching language (e.g., English) is more likely to be more accurate than its right part, while the right part is more accurate for a left-branching language (e.g., Japanese). Our discoveries are confirmed on different model structures including Transformer and RNN, and in other sequence generation tasks such as text summarization."
}

@inproceedings{ViT,
    title={An Image is Worth 16x16 Words: Transformers for Image Recognition at Scale},
    author={Alexey Dosovitskiy and Lucas Beyer and Alexander Kolesnikov and Dirk Weissenborn and Xiaohua Zhai and Thomas Unterthiner and Mostafa Dehghani and Matthias Minderer and Georg Heigold and Sylvain Gelly and Jakob Uszkoreit and Neil Houlsby},
    booktitle={ICLR},
    year={2021},
    urlX={https://openreview.net/forum?id=YicbFdNTTy}
}

@article{TUBerlin,
    author = {Eitz, Mathias and Hays, James and Alexa, Marc},
    title = {How do humans sketch objects?},
    year = {2012},
    issue_dateX = {July 2012},
    publisherX = {Association for Computing Machinery},
    addressX = {New York, NY, USA},
    volumeX = {31},
    numberX = {4},
    issnX = {0730-0301},
    urlX = {https://doi.org/10.1145/2185520.2185540},
    doiX = {10.1145/2185520.2185540},
    abstractX = {Humans have used sketching to depict our visual world since prehistoric times. Even today, sketching is possibly the only rendering technique readily available to all humans. This paper is the first large scale exploration of human sketches. We analyze the distribution of non-expert sketches of everyday objects such as 'teapot' or 'car'. We ask humans to sketch objects of a given category and gather 20,000 unique sketches evenly distributed over 250 object categories. With this dataset we perform a perceptual study and find that humans can correctly identify the object category of a sketch 73\% of the time. We compare human performance against computational recognition methods. We develop a bag-of-features sketch representation and use multi-class support vector machines, trained on our sketch dataset, to classify sketches. The resulting recognition method is able to identify unknown sketches with 56\% accuracy (chance is 0.4\%). Based on the computational model, we demonstrate an interactive sketch recognition system. We release the complete crowd-sourced dataset of sketches to the community.},
    journal = {ACM Trans. Graph.},
    monthX = jul,
    articlenoX = {44},
    numpagesX = {10},
    keywordsX = {sketch, recognition, learning, crowd-sourcing}
}

@INPROCEEDINGS{PointNet,
  author={Charles, R. Qi and Su, Hao and Kaichun, Mo and Guibas, Leonidas J.},
  booktitle={CVPR}, 
  title={PointNet: Deep Learning on Point Sets for 3D Classification and Segmentation}, 
  year={2017},
  volumeX={},
  numberX={},
  pagesX={77-85},
  keywordsX={Three-dimensional displays;Shape;Computer architecture;Feature extraction;Machine learning;Semantics},
  doiX={10.1109/CVPR.2017.16}
}

@inproceedings{BezierSketch,
    author = {Das, Ayan and Yang, Yongxin and Hospedales, Timothy and Xiang, Tao and Song, Yi-Zhe},
    title = {B\'{e}zierSketch: A Generative Model for Scalable Vector Sketches},
    year = {2020},
    isbnX = {978-3-030-58573-0},
    publisherX = {Springer-Verlag},
    addressX = {Berlin, Heidelberg},
    urlX = {https://doi.org/10.1007/978-3-030-58574-7_38},
    doiX = {10.1007/978-3-030-58574-7_38},
    booktitle = {ECCV},
    pagesX = {632–647},
    numpagesX = {16},
    keywordsX = {Sketch generation, Scalable graphics, B\'{e}zier curve},
    locationX = {Glasgow, United Kingdom}
}

@INPROCEEDINGS{SketchCloud,
  author={Das, Ayan and Yang, Yongxin and Hospedales, Timothy and Xiang, Tao and Song, Yi-Zhe},
  booktitle={CVPR}, 
  title={Cloud2Curve: Generation and Vectorization of Parametric Sketches}, 
  year={2021},
  volumeX={},
  numberX={},
  pagesX={7084-7093},
  keywordsX={Training;Graphics;Deep learning;Computer vision;Computational modeling;Fitting;Computer architecture},
  doiX={10.1109/CVPR46437.2021.00701}
}

@INPROCEEDINGS{SketchCLIP,
  author={Sain, Aneeshan and Bhunia, Ayan Kumar and Chowdhury, Pinaki Nath and Koley, Subhadeep and Xiang, Tao and Song, Yi-Zhe},
  booktitle={CVPR}, 
  title={CLIP for All Things Zero-Shot Sketch-Based Image Retrieval, Fine-Grained or Not}, 
  year={2023},
  volumeX={},
  numberX={},
  pagesX={2765-2775},
  keywordsX={Computer vision;Art;Computational modeling;Image retrieval;Performance gain;Pattern recognition;Task analysis;Vision applications and systems},
  doiX={10.1109/CVPR52729.2023.00271}
}

@article{Sketch-a-Net,
    author = {Yu, Qian and Yang, Yongxin and Liu, Feng and Song, Yi-Zhe and Xiang, Tao and Hospedales, Timothy M.},
    title = {Sketch-a-Net: A Deep Neural Network that Beats Humans},
    year = {2017},
    issue_dateX = {May       2017},
    publisherX = {Kluwer Academic Publishers},
    addressX = {USA},
    volumeX = {122},
    numberX = {3},
    issnX = {0920-5691},
    urlX = {https://doi.org/10.1007/s11263-016-0932-3},
    doiX = {10.1007/s11263-016-0932-3},
    abstractX = {We propose a deep learning approach to free-hand sketch recognition that achieves state-of-the-art performance, significantly surpassing that of humans. Our superior performance is a result of modelling and exploiting the unique characteristics of free-hand sketches, i.e., consisting of an ordered set of strokes but lacking visual cues such as colour and texture, being highly iconic and abstract, and exhibiting extremely large appearance variations due to different levels of abstraction and deformation. Specifically, our deep neural network, termed Sketch-a-Net has the following novel components: (i) we propose a network architecture designed for sketch rather than natural photo statistics. (ii) Two novel data augmentation strategies are developed which exploit the unique sketch-domain properties to modify and synthesise sketch training data at multiple abstraction levels. Based on this idea we are able to both significantly increase the volume and diversity of sketches for training, and addressX the challenge of varying levels of sketching detail commonplace in free-hand sketches. (iii) We explore different network ensemble fusion strategies, including a re-purposed joint Bayesian scheme, to further improve recognition performance. We show that state-of-the-art deep networks specifically engineered for photos of natural objects fail to perform well on sketch recognition, regardless whether they are trained using photos or sketches. Furthermore, through visualising the learned filters, we offer useful insights in to where the superior performance of our network comes from.},
    journal = {Int. J. Comput. Vision},
    monthX = may,
    pagesX = {411–425},
    numpagesX = {15},
    keywordsX = {Convolutional neural network, Data augmentation, Sketch abstraction, Sketch recognition, Stroke ordering}
}

@inproceedings{DAE,
    author = {Vincent, Pascal and Larochelle, Hugo and Bengio, Yoshua and Manzagol, Pierre-Antoine},
    title = {Extracting and composing robust features with denoising autoencoders},
    year = {2008},
    isbnX = {9781605582054},
    publisherX = {Association for Computing Machinery},
    addressX = {New York, NY, USA},
    urlX = {https://doi.org/10.1145/1390156.1390294},
    doiX = {10.1145/1390156.1390294},
    abstractX = {Previous work has shown that the difficulties in learning deep generative or discriminative models can be overcome by an initial unsupervised learning step that maps inputs to useful intermediate representations. We introduce and motivate a new training principle for unsupervised learning of a representation based on the idea of making the learned representations robust to partial corruption of the input pattern. This approach can be used to train autoencoders, and these denoising autoencoders can be stacked to initialize deep architectures. The algorithm can be motivated from a manifold learning and information theoretic perspective or from a generative model perspective. Comparative experiments clearly show the surprising advantage of corrupting the input of autoencoders on a pattern classification benchmark suite.},
    booktitle = {ICML},
    pagesX = {1096–1103},
    numpagesX = {8},
    locationX = {Helsinki, Finland},
    seriesX = {ICML '08}
}

@inproceedings{BERT,
    title = "{BERT}: Pre-training of Deep Bidirectional Transformers for Language Understanding",
    author = "Devlin, Jacob  and
      Chang, Ming-Wei  and
      Lee, Kenton  and
      Toutanova, Kristina",
    editorX = "Burstein, Jill  and
      Doran, Christy  and
      Solorio, Thamar",
    booktitle = "NAACL",
    monthX = jun,
    year = "2019",
    addressX = "Minneapolis, Minnesota",
    publisherX = "Association for Computational Linguistics",
    urlX = "https://aclanthology.org/N19-1423/",
    doiX = "10.18653/v1/N19-1423",
    pagesX = "4171--4186",
    abstractX = "We introduce a new language representation model called BERT, which stands for Bidirectional Encoder Representations from Transformers. Unlike recent language representation models (Peters et al., 2018a; Radford et al., 2018), BERT is designed to pre-train deep bidirectional representations from unlabeled text by jointly conditioning on both left and right context in all layers. As a result, the pre-trained BERT model can be fine-tuned with just one additional output layer to create state-of-the-art models for a wide range of tasks, such as question answering and language inference, without substantial task-specific architecture modifications. BERT is conceptually simple and empirically powerful. It obtains new state-of-the-art results on eleven natural language processing tasks, including pushing the GLUE score to 80.5 (7.7 point absolute improvement), MultiNLI accuracy to 86.7{\%} (4.6{\%} absolute improvement), SQuAD v1.1 question answering Test F1 to 93.2 (1.5 point absolute improvement) and SQuAD v2.0 Test F1 to 83.1 (5.1 point absolute improvement)."
}

@inproceedings{GPT,
  title={Improving language understanding by generative pre-training},
  author={Radford, Alec and Narasimhan, Karthik and Salimans, Tim and Sutskever, Ilya},
  year={2018},
  booktitle={arXiv (Preprint)},
  urlX={https://openai.com/research/language-unsupervised},
}

@INPROCEEDINGS{MAE,
  author={He, Kaiming and Chen, Xinlei and Xie, Saining and Li, Yanghao and Dollár, Piotr and Girshick, Ross},
  booktitle={CVPR}, 
  title={Masked Autoencoders Are Scalable Vision Learners}, 
  year={2022},
  volumeX={},
  numberX={},
  pagesX={15979-15988},
  keywordsX={Training;Couplings;Computer vision;Computational modeling;Computer architecture;Data models;Pattern recognition;Representation learning; Self-& semi-& meta- & unsupervised learning},
  doiX={10.1109/CVPR52688.2022.01553}
}

@inproceedings{
    StrokeCloud,
    title={Modelling complex vector drawings with stroke-clouds},
    author={Alexander Ashcroft and Ayan Das and Yulia Gryaditskaya and Zhiyu Qu and Yi-Zhe Song},
    booktitle={ICLR},
    year={2024},
    urlX={https://openreview.net/forum?id=O2jyuo89CK}
}

@inproceedings{
    PE_Extrapolation,
    title={Train Short, Test Long: Attention with Linear Biases Enables Input Length Extrapolation},
    author={Ofir Press and Noah Smith and Mike Lewis},
    booktitle={ICLR},
    year={2022},
    urlX={https://openreview.net/forum?id=R8sQPpGCv0}
}

@inproceedings{GDL,
      title={Geometric Deep Learning: Grids, Groups, Graphs, Geodesics, and Gauges}, 
      author={Michael M. Bronstein and Joan Bruna and Taco Cohen and Petar Veličković},
      year={2021},
      booktitle={arXiv (Preprint)}
}

@inproceedings{
    GSA,
    title={Group Equivariant Stand-Alone Self-Attention For Vision},
    author={David W. Romero and Jean-Baptiste Cordonnier},
    booktitle={ICLR},
    year={2021},
}

@article{SketchGNN,
    author = {Yang, Lumin and Zhuang, Jiajie and Fu, Hongbo and Wei, Xiangzhi and Zhou, Kun and Zheng, Youyi},
    title = {SketchGNN: Semantic Sketch Segmentation with Graph Neural Networks},
    year = {2021},
    journal = {ACM Trans. Graph.}
}







\end{document}